\documentclass[sigconf, nonacm]{acmart}

\AtBeginDocument{%
  }

\usepackage[ruled,vlined]{algorithm2e}

\begin{document}

\title{Compressed Concatenation of Small Embedding Models}


\author{Ben Ayad Mohamed Ayoub}
\orcid{0009-0009-0774-5412}
\authornote{Corresponding author}
\affiliation{%
  \institution{University of Passau}
  \city{Passau}
  \country{Germany}
}
\email{benaya01@ads.uni-passau.de}

\author{Michael Dinzinger}
\orcid{0009-0003-1747-5643}
\affiliation{%
  \institution{University of Passau}
  \streetaddress{Innstraße 33}
  \city{Passau}
  \country{Germany}
  \postcode{94032}
}
\email{michael.dinzinger@uni-passau.de}

\author{Kanishka Ghosh Dastidar}
\orcid{0000-0003-4171-0597}
\affiliation{%
  \institution{University of Passau}
  \streetaddress{Innstraße 33}
  \city{Passau}
  \country{Germany}
  \postcode{94032}
}
\email{kanishka.ghoshdastidar@uni-passau.de}

\author{Jelena Mitrovi\'{c}}
\orcid{0000-0003-3220-8749}
\affiliation{%
  \institution{University of Passau}
  \streetaddress{Innstraße 33}
  \city{Passau}
  \country{Germany}
  \postcode{94032}
}
\email{jelena.mitrovic@uni-passau.de}

\author{Michael Granitzer}
\orcid{0000-0003-3566-5507}
\affiliation{%
  \institution{University of Passau}
  \streetaddress{Innstraße 33}
  \city{Passau}
  \country{Germany}
  \postcode{94032}
}
\email{michael.granitzer@uni-passau.de}

\renewcommand{\shortauthors}{Ben Ayad et al.}

\begin{abstract}

Embedding models are central to dense retrieval, semantic search, and recommendation systems, but their size often makes them impractical to deploy in resource-constrained environments such as browsers or edge devices. 
While smaller embedding models offer practical advantages, they typically underperform compared to their larger counterparts.
To bridge this gap, we demonstrate that concatenating the raw embedding vectors of multiple small models can outperform a single larger baseline on standard retrieval benchmarks. 
To overcome the resulting high dimensionality of naive concatenation, we introduce a lightweight unified decoder trained with a Matryoshka Representation Learning (MRL) loss. This decoder maps the high-dimensional joint representation to a low-dimensional space, preserving most of the original performance without fine-tuning the base models.
We also show that while concatenating more base models yields diminishing gains, the robustness of the decoder's representation under compression and quantization improves. 
Our experiments show that, on a subset of MTEB retrieval tasks, our concat-encode-quantize pipeline recovers 89\% of the original performance with a 48× compression factor when the pipeline is applied to a concatenation of four small embedding models.

% add model sizes in a table

\end{abstract}






\maketitle

\section{Introduction}

Embedding models have become indispensable to information retrieval, allowing systems to match queries and documents based on semantic content rather than exact keyword overlaps. Recent progress has significantly improved the quality of these embeddings, with noticeable gains in performance on standard benchmarks like BEIR \cite{thakur2021beir} and MTEB \cite{muennighoff2022mteb}, but often at the cost of model size and complexity, making them difficult to deploy in edge environments or latency-critical applications. 

Although 33 M-parameter models such as E5 \cite{wang2022text}, GTE \cite{li2023towards}, and Arctic-Embed \cite{merrick2024arctic} ( with \texttt{snowflake-arctic-embed-s} leading the sub-100 M category as of May 17, 2025) now rival or even surpass larger counterparts on BEIR and MTEB (see the MTEB leaderboard\footnote{\url{https://huggingface.co/spaces/mteb/leaderboard}}), these smaller models still struggle to capture subtle semantic nuances compared to substantially larger architectures.

Prior work typically tackles this challenge by training increasingly larger single models from scratch
—a resource-intensive approach that ignores the complementary strengths of existing models and compounds both computational and environmental costs. 
We instead repurpose multiple \textbf{frozen} embedding models and concatenate their output vectors, producing a unified representation with broader semantic coverage. 
To address the resulting dimensionality increase, we compress the concatenated representation using a lightweight decoder trained with a Matryoshka Representation Learning (MRL) objective, preserving nearly all of the retrieval performance.

Our contributions can be summarized as follows:

\begin{itemize}
\item We demonstrate that concatenating multiple small embedding models can outperform larger single models.
\item We propose a lightweight decoder trained with MRL objective, effectively compressing high-dimensional embeddings while preserving performance.
\item We show that increasing the number of concatenated models improves the robustness of embeddings at high compression ratios. 
\end{itemize}

\section{Related work}

\paragraph{Open‑source embedding models.} Recent efforts in open-source embedding models have produced small, efficient models that narrow the performance gap with larger proprietary ones. E5 \citep{wang2022text} was the first model pre-trained with a contrastive loss \citep{chen2020simple} on a large-scale text pair dataset to beat the ranking function BM25 \cite{robertson1994some} without any fine-tuning. 
GTE's 110M model variant \cite{li2023towards} with multi-stage contrastive learning, outperforms models 10× its size on MTEB.
Arctic-Embed \cite{merrick2024arctic}, with better data sampling, hard negative mining, and improved synthetic query generation, achieved SoTA retrieval performance in many model classes, and highliting the importance of  data-centric approaches over pure scale.
BGE \cite{xiao2024c} released a series of models primarily targeting Chinese-language embeddings, widely adopted due to their ease of fine-tuning.

\paragraph{Compression for large-scale retrieval.} To enable efficient dense retrieval at scale, a range of embedding compression techniques have been explored. Classical methods, such as product quantization (PQ) \cite{jegou2010product}, and Locality Sensitive Hashing (LSH) \cite{indyk1998approximate}, 
decouple encoding from compression, and primarily optimize for reconstruction loss which can limit retrieval performance.
Recent work closes that gap by integrating quantization into training, enabling compression while better preserving retrieval accuracy and drastically reducing index size \citep{zhan2021jointly,xiao2022distill}. 
Knowledge distillation (KD) \cite{hinton2015distilling} has also been leveraged to improve embedding compression, with methods explicitly training retrievers from large language models \cite{lee2024gecko}, or recent KD-based techniques such as tightly coupled teachers, in-batch interactions, and projective distillation \cite{lin2020distilling, lin2021batch, zhao2022compressing}.

\paragraph{Adaptive representations.} A complementary line of research focuses on adaptive and multi-scale representations. Matryoshka Representation Learning (MRL) \cite{kusupati2022matryoshka} learns a single nested embedding such that truncated representations remain informative. Jina-v3 \cite{sturua2024jina} applies MRL to produce highly compressible embeddings down to 32 dimensions, with minimal quality loss. Our work builds on this paradigm by first concatenating the outputs of multiple lightweight embedding models, and then training a decoder with an MRL objective. This yields an adaptively compressible, unified representation that combines the strengths of diverse models while remaining efficient for deployment. 

%Despite rapid architectural progress in large language models, current leading embedding models—including all cited above—still predominantly rely on transformer-based architectures derived from BERT \citep{vaswani2017attention, devlin2019bert}. 
%Ensembling techniques such as concatenation or averaging are occasionally used in practice, but they typically introduce high dimensionality and inference latency, and have not been systematically studied in academic literature. 
%This work addresses that gap by concatenating the outputs of several small embedding models and learning a unified, compact representation. Importantly, the base models are kept frozen; only a lightweight decoder is trained to compress the concatenated representations into a compact embedding suitable for deployment.
% write that these methods were never exploited in the concatenated step.

\section{Methodology}
\label{sec:methodology}

\subsection{Data and architecture:}

We follow the same recipe to train all decoders. We use a corpus of $N = 500\,000$ cleaned Wikipedia passages, with each consisting of $512$ tokens. Each embedding model $S_i$ of dimensionality $d_i$ defines an embedding matrix of the original Wikipedia passages 
$E_i\in\mathbb{R}^{N\times d_i}$. Concatenating these feature matrices (w.l.o.g. for $S_1$ and $S_2$), results in the following embedding matrix
$
C=[E_1\;\|\;E_2]\in\mathbb{R}^{N\times(d_1+d_2)}
$.
This formulation yields a standard supervised‐learning problem: given the concatenated embeddings 
$
C \in \mathbb{R}^{N \times (d_1 + d_2)},
$
train a decoder
$
h : \mathbb{R}^{d_1 + d_2} \to \mathbb{R}^d,\quad d \ll d_1 + d_2,
$
to mimic the pairwise similarities of the input (raw concatenation) and output (decoder's output). For the decoder's architecture, we empirically found that a single‐layer MLP suffices—any increase in depth led to rapid overfitting.

\subsection{Training and Loss Function}
\label{sec:train}

We train
\footnote{Implementation details are available at: https://github.com/eigenAyoub/embed-fusion}
the decoder to compress concatenated embeddings into a lower‐dimensional space while preserving \emph{cosine} pairwise similarities.  Let 
$Z = \{z_j\}_{j=1}^B \subset \mathbb{R}^{d_1 + d_2},$
be a batch of size \(B\), and let 
%$H = h(Z) \in \mathbb{R}^{B \times d}, \quad d \ll d_1 + d_2.$
$H = h(Z) \in \mathbb{R}^{B \times d},$
be the decoder's output, with $h_i$ the $i$-th row of $H$.
We define the similarity loss for a single batch 
%for dimension $d$ (full dimensionality) 
as follows: 
\[
\ell_{\mathrm{sim}}(H,\,Z)
= \frac{1}{B(B-1)} \sum_{i\neq j}
\Bigl[\cos(h_i, h_j) - \cos(z_i, z_j)\Bigr]^2,
\]
%where \(\cos(a,b)=\tfrac{\langle a,b\rangle}{\|a\|\|b\|}\), 

The overall loss that we backpropagate through is defined as the average of the similarity loss ($\ell_{\mathrm{sim}}$) with varying truncated dimensions of the decoder's outputs.
We truncate following a set of Matryoshka dimensions (or stops), \(\mathcal{D}=\{d^{(1)},\dots,d^{(K)}\}\):

\[
\mathcal{L}
= \frac{1}{\lvert \mathcal{D} \rvert}
\sum_{i \in \{1, \dots, K\}}
\ell_{\mathrm{sim}}\bigl(H{[:,:\,d^{(i)}]},\,Z\bigr).
\]

where \(H{[:,:\,j]}\) denotes the first \(j\) columns of \(H\). Each stop explicitly encourage the decoder to enhance its representation at it.
%focus on the predefined set of dimensions $\mathcal{D}$. 
\paragraph{Decoder's output dimension.} Unless stated as a subscript (e.g., \(\operatorname{Dec}_{1024}\)), all decoders produce a 768‐dim output, and trained with the following MRL stops:
$\{32,\;64,\;128,\;200,\;256,\;300,\;384,\;512,\;768\}.$

\paragraph{Overall size.}  The largest combination \(M_1S_i\) totals 142 M parameters (109M for $M_1$, and 33M $S_1$). The decoder adds a negligible size \(\bigl(1152\times768\approx0.9\text{ M}\bigr)\), which brings the total size to \(\approx142.9\text{ M}\).

%The decoder adds a negligible number of parameters.

\subsection{Quantization}
\label{sec:quantization}

Quantization has two stages.  
\textbf{Offline calibration} learns per-dimension break-points on a reference set;  
\textbf{online inference} reuses those break-points to map new decoder outputs to \(b\)-bit codes.

\paragraph{Offline calibration.}
Draw \(S=100\,000\) Wikipedia passages, form their concatenated embeddings
$Z_{\mathrm{ref}}\in\mathbb{R}^{S\times(d_{1}+d_{2})}$, and encode them with the best decoder, 
$
H_{\mathrm{ref}} = h(Z_{\mathrm{ref}})\in\mathbb{R}^{S\times d}.
$ For each output dimension \(j=1,\dots,d\) compute the \((2^{b}-1)\) empirical
percentiles
\[
\tau_{j,k}
\;=\;
\operatorname{Percentile}\!\bigl(
        H_{\mathrm{ref},:,j},\,
        100\,k/2^{b}
      \bigr),
\quad
k=1,\dots,2^{b}-1.
\]
Define \(\tau_{j,0}=\min H_{\mathrm{ref},:,j}\) and
\(\tau_{j,2^{b}}=\max H_{\mathrm{ref},:,j}\).
The set \(\{\tau_{j,k}\}_{k=0}^{2^{b}}\) partitions the axis into \(2^{b}\)
equal-mass buckets.

\paragraph{Online inference.}
Given any new batch of embeddings
$Z' \!\in\! \mathbb{R}^{N\times(d_{1}+d_{2})}$,
we first evaluate the decoder,
$
H' \;=\; h(Z') \in \mathbb{R}^{N\times d},
$
and then assign a $b$-bit symbol to every coordinate via
\[
q_{i,j}
   \;=\;
   \sum_{k=1}^{2^{b}-1}
   \bigl[\,H'_{i,j}>\tau_{j,k}\bigr],
   \qquad
   i=1,\dots,N,\;
   j=1,\dots,d.
\]

%The resulting code matrix is
%\(Q\in\{0,\dots,2^{b}-1\}^{N\times d}\),
%obtained without any further calibration.

\subsection{Retrieval Evaluation}
\label{sec:evaluation}

We evaluate our embedding models on downstream retrieval using the Massive Text Embedding Benchmark (MTEB) \cite{muennighoff2022mteb}, focusing on a subset of the BEIR \cite{thakur2021beir}. We assess our models on six heterogeneous tasks: NFCorpus (clinical notes to consumer FAQs), SciFact (claim verification in scientific abstracts), ArguAna (argument retrieval in online debates), SciDocs (citation and co-citation recommendation), AILAStatutes (legal domain), and QuoraRetrieval (duplicate question retrieval). Spanning biomedical, scientific, legal, and open-domain question–answer settings, and involving both short (claims, questions) and long-form (abstracts, posts) text pairs. We report Normalized Discounted Cumulative Gain at rank 10 (nDCG@10). 
%following the recommendations of BEIR \cite{thakur2021beir}.

\section{Raw concatenation}

We evaluate the raw concatenation of several pairs of embedding models that vary in size and output dimensionality, as listed in Table \ref{tab:dims}. Table \ref{tab:pairs} shows that, in most cases, concatenating two models boosts performance over each participant model and occasionally surpasses much larger baselines.
For example, concatenating \texttt{Arctic-m} with \texttt{bge-small} ($[M_{1},S_{5}]$) and testing on all retrieval tasks in the MTEB benchmark\footnote{\url{https://huggingface.co/PaDaS-Lab/arctic-m-bge-small}} yields a mean score of 56.5. At the time of release, this placed the model 32nd on the (legacy) MTEB leaderboard, outperforming substantially larger models such as \texttt{gte-Qwen2-7B}\footnote{\url{https://huggingface.co/Alibaba-NLP/gte-Qwen2-7B-instruct}} \cite{li2023towards}, which scored 56.24.

\begin{table*}[h]
    \centering
    \caption{Embedding models: parameter counts and output dimensions}
    \label{tab:dims}
    \begin{tabular}{@{}lccccc@{}}
        \toprule
        \textbf{S$_i$} & \textbf{M$_i$} & \textbf{[M$_i$,S$_j$]} & \textbf{[S$_i$,S$_j$]} & \textbf{T$_1$} \\
        \midrule
        33M & 109M & 142M & 66M & 335M \\
        384 & 768 & 1152 & 768 & 1024\\
        \bottomrule
    \end{tabular}
\end{table*}

\begin{table*}[t]
    \centering
    \caption{Retrieval performance of selected pairs of embedding models}
    \label{tab:pairs}
    \begin{tabular}{@{}lccccccc@{}}
        \toprule
                            & \textbf{NFCorpus} & \textbf{SciFact} & \textbf{ArguAna} & \textbf{SciDocs} & \textbf{AILAStatutes} & \textbf{QuoraRetrieval} 
                            &\textbf{average}
                            \\ 
        \midrule
    \textbf{e5-small ($S_1$)}    & 0.32461 &  0.68750 & 0.41794   & 0.17714 & 0.20214 & 0.84946  & 0.44313  \\
    \textbf{No-Ins ($S_2$)}      & 0.34921 & 0.72219 & 0.57593   & 0.21823 & 0.28411 & 0.88413   & 0.50563  \\
    \textbf{gte-small ($S_3$)}   & 0.34767 & 0.72701  &  0.55423 & 0.21394 & 0.24066 & 0.88017   & 0.49395  \\
    \textbf{bge-small ($S_5$)}   & 0.33708 & 0.72000 & 0.59499   & 0.19725 & 0.20813 & 0.88783   & 0.49088  \\
    \textbf{Arctic-m ($M_1$)}    & 0.36236 & 0.71586 & 0.59530   & 0.21492 & 0.28101 & 0.87366   & 0.50718  \\
        \midrule
%        \midrule
    \textbf{$[M_{1},S_{3}]$} & 0.37305 & 0.73927 & 0.60039 & 0.22641 & 0.28166 & 0.88625 & 0.51784 \\
   \textbf{Dec([$M_{1},S_{3}$])[:384]} & 0.36514 & 0.72306 & 0.59297 & 0.21870 & 0.27719 & 0.87817 & 0.50920 \\
        \midrule
   \textbf{[$M_{1},S_{5}$]} & 0.37561 & 0.73427 & 0.62660 & 0.22672 & 0.26633 & 0.89245  & 0.52033 \\
    \textbf{Dec([$M_{1},S_{5}$])[:384]} & 0.37054 & 0.72834 & 0.62268 & 0.21998 & 0.27700 & 0.88854 & 0.51780 \\
        \midrule
    \textbf{$T_1[:256]$} & 0.36213 & 0.69321 & 0.62297 & 0.21489 & 0.24152 & 0.88244 & 0.50286 \\
    \textbf{Dec([$M_{1},S_{5}$])[:256]} & 0.36237 & 0.71579 & 0.61370 & 0.21570 & 0.26388 & 0.88290 & 0.50906 \\
        \midrule
    \textbf{$[T_1[:384],M_1[:384]]$} & 0.37760  & 0.71351  & 0.64377  & 0.22154  & 0.28586 & 0.88542  & 0.52128 \\
    \textbf{$\textbf{Dec}([T_1,M_1])[:384]$} & 0.38022 & 0.74014 & 0.64918 & 0.23053 & 0.24610 & 0.88641 & 0.52210 \\
        \midrule
        %——— all pairs ——%
        \textbf{[S$_1$, S$_2$]} ($C_2$)    & 0.35753 & 0.73271 & 0.56407 & 0.22079 & 0.26426 & 0.88572 & 0.50418 \\
        \textbf{Dec([$S_{1},S_{2}$])[:384]} & 0.35265 & 0.73345 & 0.55236 & 0.21454 & 0.25383 & 0.88350 & 0.49839 \\
        \midrule
        \textbf{[S$_1$, S$_5$]}    & 0.35594 & 0.72684 & 0.58599 & 0.20275 & 0.22882 & 0.88843 & 0.49813 \\
        \textbf{Dec([$S_{1},S_{5}$])[:384]} & 0.34703 & 0.72602 & 0.58407 & 0.20171 & 0.24530 & 0.88870 & 0.49881 \\
        \midrule
        \textbf{[S$_3$, S$_5$]}    & 0.35566 & 0.73310 & 0.59653 & 0.21121 & 0.22114 & 0.88749 & 0.50086 \\
        \textbf{Dec$_{1024}([S_{3},S_{5}])[:512]$} & 0.35207 & 0.73206 & 0.59904 & 0.21001 & 0.23302 & 0.88736 & 0.50226 \\
        \textbf{Dec$_{512}([S_{3},S_{5}])$} & 0.35306 & 0.72698 & 0.59563 & 0.21152 & 0.23808 & 0.88709 & 0.50206 \\
        \bottomrule
    \end{tabular}
\end{table*}

\paragraph{Model Selection.} We find that concatenating models with complementary strength (performing well on different tasks) yields stronger results, suggesting that diversity across models enhances robustness. However, performance gains diminish with scale as the improvement from two to four models (C1 vs. C2) is less pronounced. We also, noticed that heavily fine-tuned models perform poorly when concatenated, or compressed with a decoder. Hence why we prefer combining base models ($S_1$, $S_3$, and $S_5$).

%A larger hybrid, $[M_{1},S_{1},S_{2},S_{3}]$, combines 209 M parameters yet delivers a sizeable boost over every constituent model and comes within striking distance of \texttt{MXBAI} ($T_{1}$), a 335 M-parameter baseline.

\section{Effectiveness of the compression}

To mitigate the linear growth in dimensionality of naive concatenation, we train 
(following the process in Methodology \ref{sec:methodology}) a decoder to map the representation of a concatenation of small embedding models to a lower dimension, and show that it performs as good the original concatenated representation, and in some cases outperforming it. 
To show the effectiveness of the decoder we first evaluate it on single models and then pairs of models.  

\subsection{Single Model Evaluation}

We show that training a decoder on the output of single models improves retrieval accuracy even when those models were originally trained with an MRL objective. Specifically, we evaluate, \texttt{Arctic-m} (denoted $M_{1}$) and \texttt{MXBAI} ($T_{1}$).
As Table~\ref{tab:single} shows, truncating to the first 384 dimensions ($M_{1}[:384]$ and $T_{1}[:384]$) drops the performance when compared to the full dimensional representations.  
When we instead train a decoder on the full representations (denoted $\mathrm{Dec}(M_{1})$ and $\mathrm{Dec}(T_{1})$) and then evaluate the decoder’s first 384 dimensions ($\mathrm{Dec}(M_{1})[:384]$, $\mathrm{Dec}(T_{1})[:384]$), both models not only recover but improve upon their truncated baseline. These findings align with \cite{yoon2024matryoshka}, but unlike their work, we target small models, and train each decoder without any task-specific corpus.

\begin{table*}[t]
    \centering
    \caption{Retrieval performance of selected single models}
    \label{tab:single}
    \begin{tabular}{@{}lccccccc@{}}
        \toprule
                            & \textbf{NFCorpus} & \textbf{SciFact} & \textbf{ArguAna} & \textbf{SciDocs} & \textbf{AILAStatutes} & \textbf{QuoraRetrieval} &\textbf{Average}\\ 
        \midrule
    \textbf{mxbai ($T_1$)}  & 0.38643 & 0.73893 & 0.65467 & 0.23101 & 0.24754 & 0.88847 & 0.52451 \\
    \textbf{Arctic-m ($M_1$)}    & 0.36236 & 0.71586 & 0.59530   & 0.21492 & 0.28101 & 0.87366   & 0.50718  \\
        \midrule
\textbf{$T_1[:384]$}        & 0.37769  & 0.71114  & 0.64241  & 0.22143  &  0.24768 & 0.88533  & 0.51428 \\
\textbf{Dec($T_1$)[:384]}   & 0.38078  & 0.73622  & 0.64935  & 0.22963  & 0.24218 & 0.88627  & 0.52074  \\
        \midrule
\textbf{$M_1[:384]$}        & 0.35242  & 0.70265  & 0.58932  & 0.21050  & 0.29594 & 0.86179  & 0.50210 \\
\textbf{Dec($M_1$)[:384]}   & 0.36394  & 0.71668  & 0.59132  & 0.21343  & 0.27720 & 0.86608  & 0.50477 \\
        \bottomrule
    \end{tabular}
\end{table*}

\subsection{Pairwise Model Evaluation}

When we train a decoder on the concatenation of pairs of models, we can in most cases recover up to 98\% of the performance of the original concatenation for a target output of 384-d, regardless of the input dimensionality, as highlighted next from Table~\ref{tab:pairs}:
%(i.e., the number of concatenated models). We highlight three examples from

\begin{itemize}
    \item \textbf{Dec([$M_{1},S_{5}$])[:384]} outperforms the full representation of $M_1$ with only half the output dimensionality, and improves over all other 384-dimensional representations above it.
    
    \item \textbf{Dec([$M_{1},S_{5}$])[:256]} outperforms $T_1[:256]$ using only 143M parameters, with \textbf{$T_1$} being a 335M parameters model.
    
    \item \textbf{Dec([$T_{1},M_{1}$])[:384]} improves over all 384-d representations, including $M_1[:384]$, $T[:384]$, and their concatenation $[T_1[:384], M_1[:384]]$. $T_1$ and $M_1$ were both trained with MRL. 
\end{itemize}

\subsection{Effect of the decoder's size, and MRL stops.}

Following the terminology introduced in Section~\ref{sec:train}, we train two decoders, $\operatorname{Dec}_{512}$, and $\operatorname{Dec}_{1024}$, on the concatenated features $[S_3, S_5]\in\mathbb{R}^{768}$. For $\operatorname{Dec}_{512}:\mathbb{R}^{768}\to\mathbb{R}^{512}$, we use the following set of Matryoshka stops: $\mathcal{D}_{512} = \{384, 512\}$. For $\operatorname{Dec}_{1024}:\mathbb{R}^{768}\to\mathbb{R}^{1024}$, we use 
$
\mathcal{D}_{1024} = \{32,\,64,\,128,\,256,\,384,\,512,\,768,\,892,\,1024\},
$
and then truncate $\operatorname{Dec}_{1024}$ to its first 512 dimensions. The results presented in the last section of Table \ref{tab:pairs}, highlight a general trend: decoders trained with a wide output (1024) and larger set of Matryoshka stops, then truncated to a target dimension (512) outperform decoders that are trained directly on the target dimension.

\subsection{Robustness Under Extreme Compression and Quantization}

\begin{table*}[t]
    \centering
    \caption{Robustness of the concatenation under extreme compression ratios}
    \label{tab:robust}
    \begin{tabular}{@{}lccccccc@{}}   % 1 (l) + 7 (c) = 8 columns
        \toprule
                            & \textbf{NFCorpus} & \textbf{SciFact} & \textbf{ArguAna} & \textbf{SciDocs} & \textbf{AILAStatutes} & \textbf{QuoraRetrieval} & \textbf{Average} \\ 
        \midrule
        \textbf{$[S_1,S_2,S_3,S_5]$ ($C_1$)}  & \textbf{0.35983} & \textbf{0.73603} & \textbf{0.59773} & 0.21912 & \textbf{0.24400} & \textbf{0.88891} & \textbf{0.50760} \\
        \textbf{$[S_1,S_2]$ ($C_2$)}         & 0.35753          & 0.73271          & 0.56407          & \textbf{0.22079} & 0.26426          & 0.88572          & 0.50418 \\
        \midrule
        \textbf{$\mathrm{LSH}_{8192}(C_1)$}   & 0.35284          & 0.71909          & \textbf{0.57986} & 0.21005          & \textbf{0.23938} & \textbf{0.88539} & 0.49777 \\
        \textbf{$\mathrm{LSH}_{8192}(C_2)$}   & \textbf{0.35333} & \textbf{0.72860} & 0.54454          & \textbf{0.21362} & 0.20858          & 0.88118          & 0.48831 \\
        \midrule
        \textbf{$\mathrm{LSH}_{1024}(C_1)$}   & \textbf{0.32840} & \textbf{0.67087} & \textbf{0.52364} & \textbf{0.19328} & \textbf{0.24645} & \textbf{0.87240} & \textbf{0.47251} \\
        \textbf{$\mathrm{LSH}_{1024}(C_2)$}   & 0.32218          & 0.66514          & 0.47661          & 0.18697          & 0.16807          & 0.86303          & 0.44700 \\
        \bottomrule
    \end{tabular}
\end{table*}
While simply concatenating more embedding models yields diminishing returns, we find that it improves robustness of the output representation under extreme compression and quantization scenarios. As shown in Table~\ref{tab:robust}, the two combinations \(C_{1}\) (132M parameters) and \(C_{2}\) (66M parameters) remain relatively competitive in retrieval performance. Applying a random projection to a high dimensional space (8192 dimensions) then 1‐bit quantization to \(C_{1}\) (\(\mathrm{LSH}_{8192}(C_{1})\)), produces comparable results for both combinations, with the smaller combination \(C_{2}\) even outperforming $C_1$ on three out of six tasks. However, in extremely compressed scenarios--- such as projecting down to 1024 dimensions (a 48× compression factor for $C_1$) \(\mathrm{LSH}_{1024}(C_{1})\) outperforms \(\mathrm{LSH}_{1024}(C_{2})\) across all tasks.

\section{Conclusion}

Embedding model deployment in resource-constrained environments remains challenging due to trade-offs between performance and practicality. To address this, we concatenate multiple small embedding models and compress the resulting high-dimensional representation using a lightweight decoder trained via a Matryoshka Representation Learning objective. Our method recovers most of the performance of the original concatenated embeddings while significantly reducing dimensionality and enhancing robustness under aggressive compression and quantization. This approach offers an efficient and practical alternative to training large models from scratch.

\begin{acks}
This project has received funding from the European Union's Horizon Europe research and innovation programme under the \href{https://cordis.europa.eu/project/id/101070014}{OpenWebSearch.EU project}.
\end{acks}

\clearpage
\section*{GenAI Usage Disclosure}

The authors used generative AI tools to assist with initial code development and minor text revisions.
Specifically, AI models helped generate early versions of some code segments, which the authors carefully reviewed, modified, and adapted into the code base. Generative AI was also used for minor editorial improvements to enhance clarity and readability. The authors have verified all AI-assisted content and take full responsibility for its accuracy and integrity.

\bibliographystyle{ACM-Reference-Format}
\bibliography{sample-base}

\end{document}